\documentclass[runningheads]{llncs}
\usepackage{cite}
\usepackage{amsmath,amssymb,amsfonts}
\usepackage{algorithmic}
\usepackage{graphicx}
\usepackage{textcomp}
\usepackage{xcolor}
\usepackage{multicol}
\usepackage{multirow}
\usepackage{dblfloatfix} 
\usepackage{float}
\usepackage{subfig}  
\usepackage{amsmath}

\begin{document}
\title{An Improvement for Capsule Networks using Depthwise Separable Convolution}
%
%
\author{Nguyen Huu Phong* \and
Bernardete Ribeiro}
\authorrunning{Phong N.H. and Bernardete R.}
\titlerunning{An Improvement for Capsule Networks...}

%
\institute{CISUC, Department of Informatics Engineering, University of Coimbra, Portugal
\email{\{phong,bribeiro\}@dei.uc.pt}}

\maketitle              
\begin{abstract}
Capsule Networks face a critical problem in computer vision in the sense that the image background can challenge its performance, although they learn very well on training data. 
In this work, we propose to improve Capsule Networks' architecture by replacing the Standard Convolution with a Depthwise Separable Convolution. This new design significantly reduces the model's total parameters while increases stability and offers competitive accuracy. In addition, the proposed model on $64\times64$ pixel images outperforms standard models on $32\times32$ and $64\times64$ pixel images. Moreover, we empirically evaluate these models with Deep Learning architectures using state-of-the-art Transfer Learning networks such as Inception V3 and MobileNet V1. The results show that Capsule Networks can perform comparably against Deep Learning models. To the best of our knowledge, we believe that this is the first work on the integration of Depthwise Separable Convolution into Capsule Networks.
\keywords{Capsule Networks \and Depthwise Separable Convolution \and  \\ Deep Learning \and Transfer Learning.}
\end{abstract}

\section{Introduction}
\label{sec:Introduction}
In our previous research, we performed experiments to compare accuracy and speed of Capsule Networks versus Deep Learning models. We found that even though Capsule Networks have a fewer number of layers than the best Deep Learning model using MobileNet V1 \cite{howard2017mobilenets}, the network performs just slightly faster than its counterpart.

We first explored details of MobileNet V1's architecture and observed that the model utilizes Depthwise Separable Convolution for the speed improvement. The layer comprises of a Depthwise Convolution and Pointwise Convolution which sufficiently reduces the model size and computation. Since Capsule Networks also integrate a Convolution layer in its architecture, we propose to replace this layer with the faster Convolution.

At the time this article is being written, there are $439$ articles citing the original Capsule Networks paper \cite{sabour2017dynamic}. Among these articles, a couple of works attempt to improve speed and accuracy of Capsule Networks. For example, the authors \cite{bahadori2018spectral} propose Spectral Capsule Networks which is composed by a voting mechanism based on the alignment of extracted features into a one-dimensional vector space. In addition, other authors claim that by using a Convolutional Decoder in the Reconstruction layer could decrease the restoration error and increase the classification accuracy \cite{mobiny2018fast}.

Capsule Networks illustrated its effectiveness on MNIST dataset, though much variations of background to models e.g. in CIFAR-10 probably causes the poorer performance \cite{sabour2017dynamic}. To solve this problem, we argue that primary filters to eliminate such backgrounds should be as important as other parts of Capsule Networks' architecture. Just, improvements on speed and accuracy of these Convolution layers can be bonus points for the network.

After a thorough search of relevant literature, we believe that this is the first work on the integration of Depthwise Separable Convolution into Capsule Networks.

The rest of this article is organized as follows. In Section~\ref{sec:contrib}, we highlight our main contributions. Next, we analyse the proposed design in Section~\ref{sec:capsule}. Following,  the design of Deep Learning models for the purpose of comparison is illustrated in Section ~\ref{sec:deeplearning}. Experiments and results are discussed in Section ~\ref{sec:experiments} accordingly. Finally, we conclude this work in Section~\ref{sec:conclude}.
\section{Contribution}
\label{sec:contrib}
Our main contributions are twofold: first, on the replacement of Standard Convolution with Depthwise Separable Convolution in Capsule Networks' Architecture and, second, on the empirical evaluations of the proposed Capsule Networks against Deep Learning models.

Regarding the design of Capsule Networks, we found that the integration of Depthwise Separable Convolution can significantly reduce the model size, increase stability and yield higher accuracy than its counterpart.

With respect to experimental evaluations of Capsule Networks versus Deep Learning models, the Capsule models perform competitively both on model size and accuracy.
\section{Integration of Depthwise Separable Convolution and Capsule Networks}
\label{sec:capsule}
\begin{figure*}[!t]
\begin{center}
\includegraphics[width=0.95\textwidth]{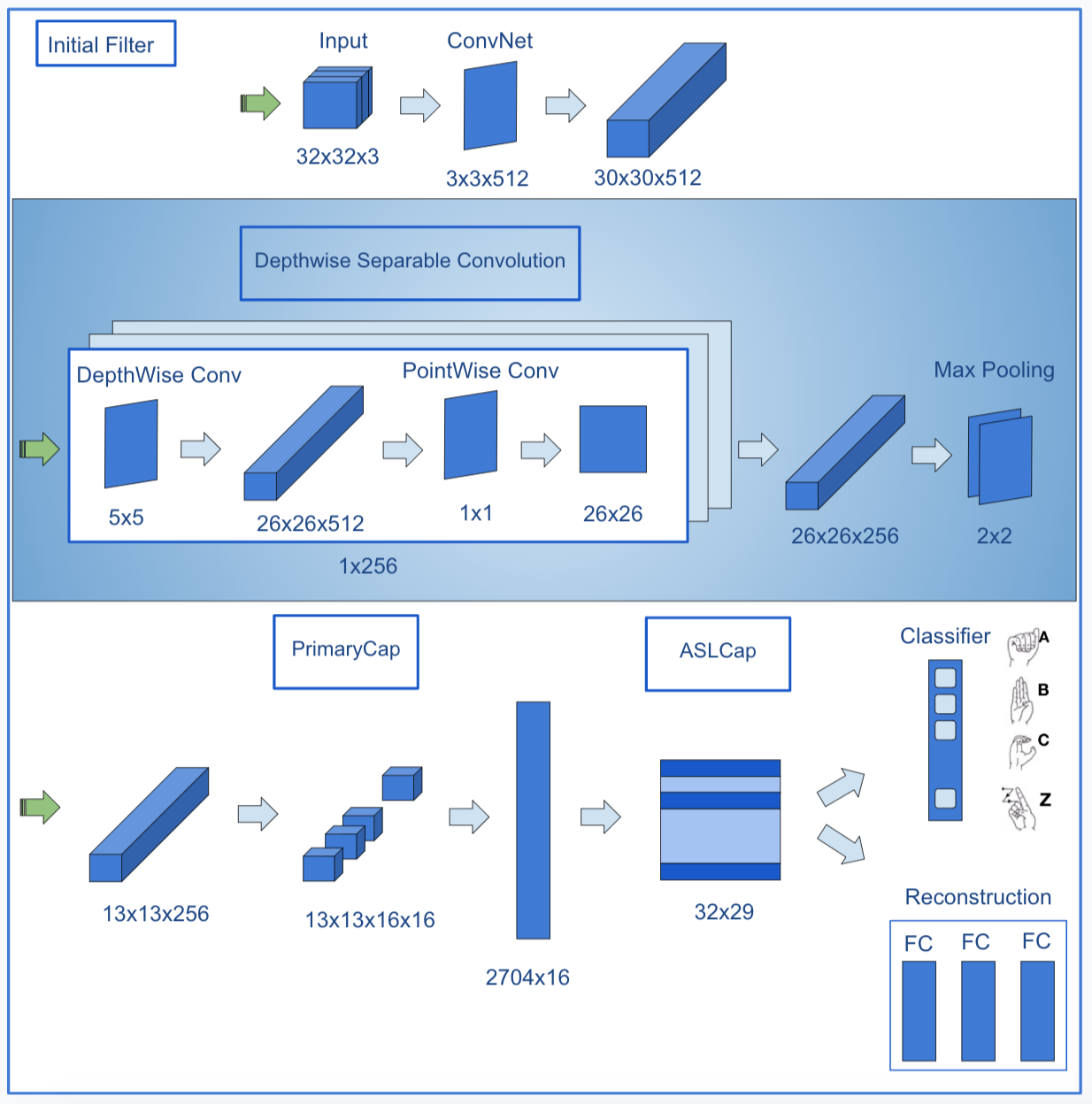}
\caption{Depthwise Separable Convolution Capsule Architecture.}
\label{fig:asl_capsule}
\end{center}
\end{figure*}
As mentioned in the previous Section, we propose to substitute a Standard Convolution (SC) of Capsule Networks with a Depthwise Separable Convolution (DW). Figure \ref{fig:asl_capsule} illustrates the architecture of the proposed model. Additionally, we apply this architecture for an American Sign Language (ASL) dataset with $29$ signs. We discuss about this dataset more details in Section ~\ref{sec:dataset}. 

We divide this architecture into three main layers including Initial Filter, Depthwise Separable Convolution Layer and Capsules Layer. In principle, we can change the first Standard Convolution with a Depth Separable Convolution. However, the Input images have only three depth channels which require fewer computations than in the second layer so that we keep the first Convolution intact. The reduction of computation cost using DW is formulated as follows.

An SC takes an input $D_{F}\times D_{F}\times M$ feature map {F} and generates $D_{G}\times D_{G}\times N$ feature map {G} where $D_{F}$ is the width and height of the input, $M$ is the number of input channels, $D_{G}$ is the width and height of the output, and $N$ is the number of output channels.

The SC is equipped with a kernel of size $D_{K}\times D_{K}\times M \times N$ where $D_{K}$ is assumed to be a spacial square. $M$ and $N$ are the number of input and output channels as mentioned above.

The output of the feature map {G} using an SC with kernel stride is 1 and same padding (or padding in short):
\begin{equation}
G_{k,l,n} = \sum_{i,j,m} K_{i,j,m,n} \cdot F_{k+i-1, l+j-1,m}
\end{equation}

The computation cost of the SC can be written as:
\begin{equation}
D_{K}\cdot D_{K} \cdot M \cdot N \cdot D_{F} \cdot D_{F}
\end{equation}

For one channel, the output of feature map \^{G} after Depthwise Convolution:
\begin{equation}
\hat{G}_{k,l,m} = \sum_{i,j,m} \hat{K}_{i,j,m} \cdot F_{k+i-1, l+j-1,m}
\end{equation}
where \^{K} is the denotation of a Depthwise Convolution with kernel size $D_{K}\times D_{K} \times M$ and this Convolution applies $m_{th}$ filter to $m_{th}$ channel of {F} yields the $m_{th}$ channel in the output.

The computation cost of the Depthwise Convolution is written as:
\begin{equation}
D_{K}\cdot D_{K} \cdot M \cdot D_{F} \cdot D_{F}
\end{equation}

Pointwise Convolution uses $1\times1$ Convolution, therefore, the total cost after Pointwise Convolution:
\begin{equation}
D_{K}\cdot D_{K} \cdot M \cdot D_{F} \cdot D_{F} + M \cdot N \cdot D_{F} \cdot D_{F}
\end{equation}

The computation cost of Depthwise Separable Convolution after two convolutions is reduced as:
\begin{equation}
\begin{split}
\frac{D_{K}\cdot D_{K} \cdot M \cdot D_{F} \cdot D_{F} + M \cdot N \cdot D_{F} \cdot D_{F}}{D_{K}\cdot D_{K} \cdot M \cdot N \cdot D_{F} \cdot D_{F}}
\\
=\frac{1}{N}+\frac{1}{D_{K}^2}
\end{split}
\end{equation}

For more details of these computations, please refer to the MobileNet V1 article \cite{sabour2017dynamic}.

In Figure \ref{fig:asl_capsule}, we show a sample where colour images have a size of $32\times32\times3$ and filtered by a ConvNet which is equipped with a $3\times3$ kernel and $512$ filters. If the DW Conv is applied in this layer, then the cost of computation can be reduced between $8$ and $9$ times. Though the input image has a depth of only $3$, the reduction of computation cost in this layer is unmatched with hundreds of channels in the second layer.


%
\section{Deep Learning Models}
\label{sec:deeplearning}
\begin{figure}[h]
\begin{center}
\includegraphics[width=0.95\textwidth]{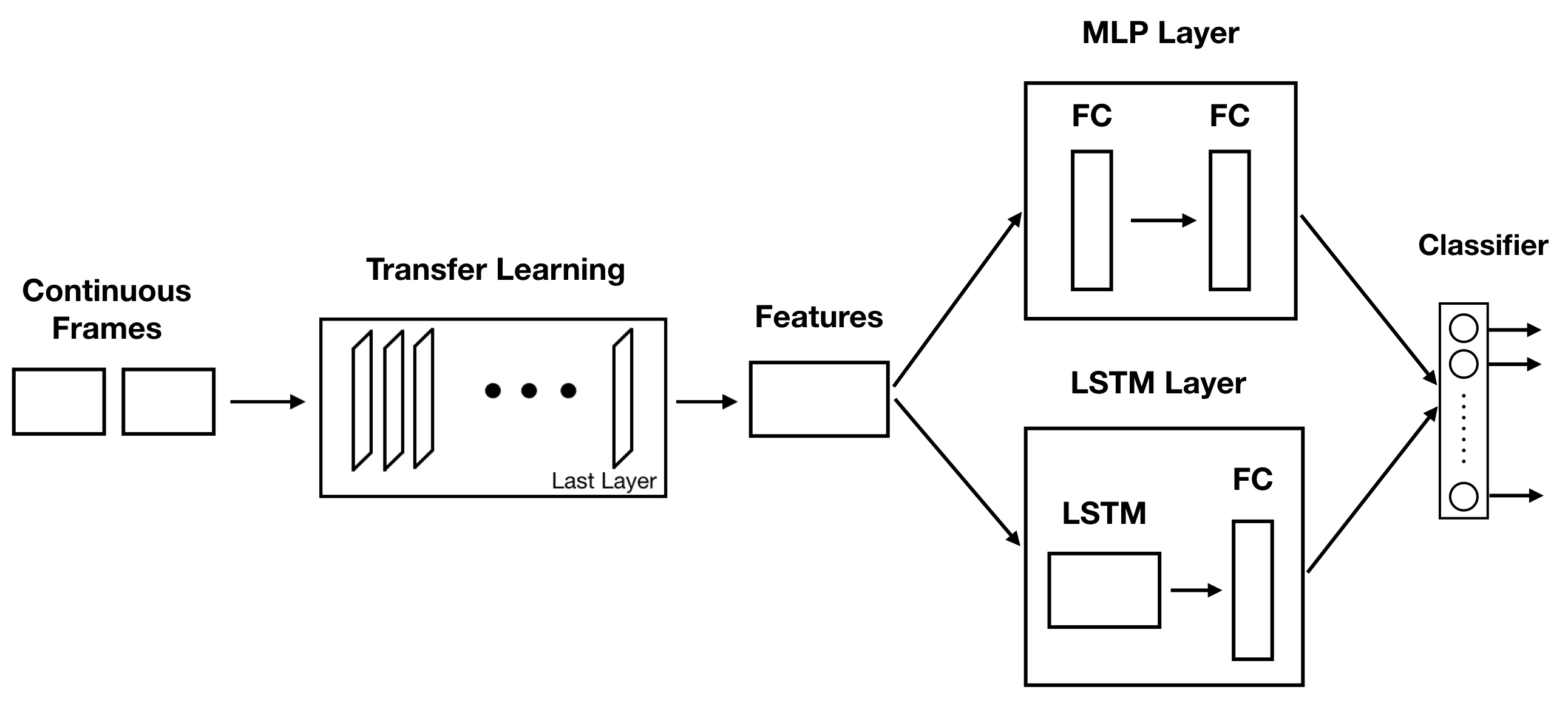}
\caption{Deep Learning Architecture}
\label{fig:asl_deeplearning}
\end{center}
\end{figure}
For the purpose of comparison, we also briefly discuss the Deep Learning's architecture as shown in Figure \ref{fig:asl_deeplearning}. As we can see from the Figure, this architecture comprises of several crucial layers including Transfer Learning, Multilayer Perceptron (MLP) and Long Short Term Memory (LSTM) layers. In Transfer Learning layer, we utilize one of the models including Inception V3 \cite{szegedy2016rethinking}, DenseNet V201 \cite{huang2017densely}, NASNetMobile \cite{zoph2018learning}, MobileNet V1 \cite{howard2017mobilenets} and MobileNet V2 \cite{sandler2018mobilenetv2} to extract features from input ASL signs. We then use MLP Layer as a baseline to compare with LSTM Layer. The MLP Layer includes two Fully Connected (FC) Neural Networks each with $512$ neurons whereas the LSTM Layer contains one LSTM with $2048$ units and one FC. More detail of the architecture can be referred to our earlier work.

\section{Experiments and Results}
\label{sec:experiments}
In this section, we first discuss an ASL Dataset that will be used as our testbed. Then we setup experiments to compare our proposed Capsule's architecture using Depthwise Separable Convolution against typical Convolution Networks. Next, we analyse performances of Transfer Learning models including Inception V3, DenseNet V201, NASNet, MobileNet V1 and MobileNet V2 using MLP and LSTM. Finally, we pickup the best of Capsule Networks and challenge the best of Deep Learning models. 
\subsection{ASL Dataset}
\label{sec:dataset}
\begin{figure}[!t]
\begin{center}
\includegraphics[width=0.95\textwidth]{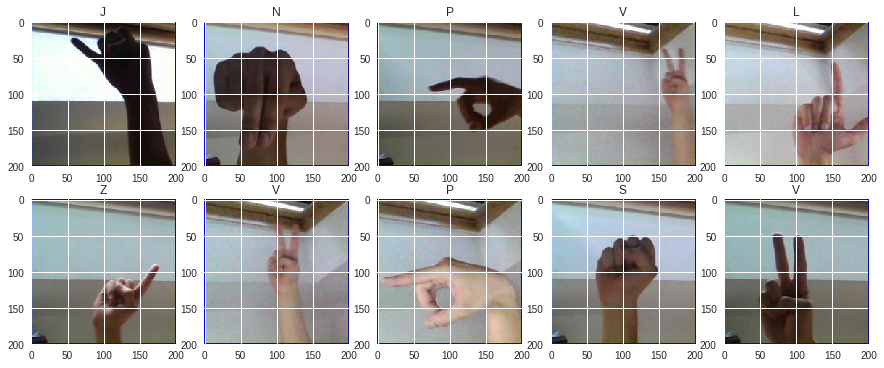}
\caption{Random Samples from the ASL Dataset}
\label{fig:ASL_Dataset}
\end{center}
\end{figure}

For the purpose of comparison with our previous work, we use the same ASL dataset for fingerspelling. The dataset was obtained from Kaggle website and includes 26 signs for letters A to Z with 3 additional signs using in other cases. Each sign contains $3000$ samples ($200 \times 200$ pixels), totally $87000$ samples for all signs. Figure~\ref{fig:ASL_Dataset} shows 10 random samples from this dataset. In these experiments, we use half of the number of samples since the accuracy are similar to that of using all dataset and this also reduces the training time by half. We divide the data into a train set and a test set with the ratio $70$ and $30$.
\subsection{Experiment 1: DW Capsules vs SC Capsules}
\begin{figure*}[p]
  \centering
  \subfloat[DW vs SC V1]{\includegraphics[width=0.475\textwidth]{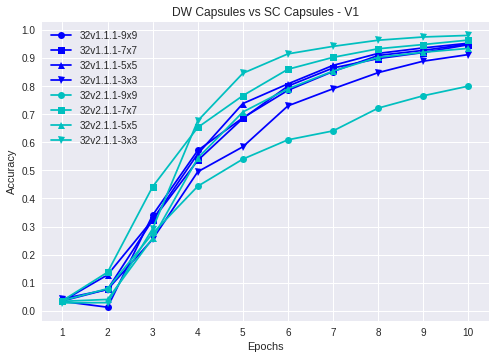}
  \label{fig:dw_vs_sc_a}}
  \subfloat[DW vs SC V2]{\includegraphics[width=0.475\textwidth]{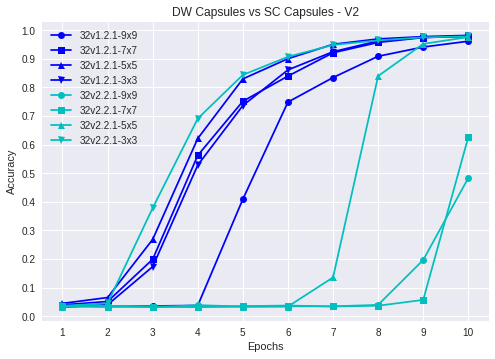}
  \label{fig:dw_vs_sc_b}}
  \\
  \subfloat[DW vs SC Mini]{\includegraphics[width=0.475\textwidth]{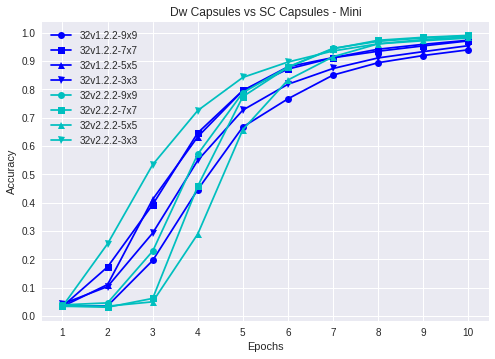}
  \label{fig:dw_vs_sc_c}}
  \subfloat[DW vs SC Max]{\includegraphics[width=0.475\textwidth]{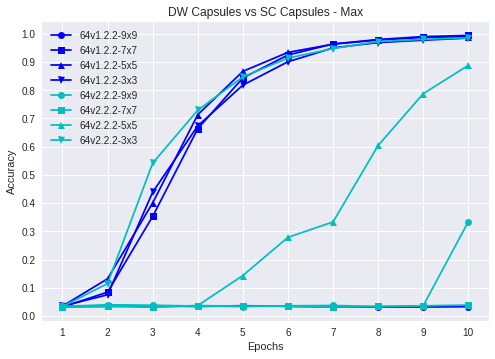}
  \label{fig:dw_vs_sc_d}}
  \\  
  \caption{DW Capsules (blue) vs SC Capsules (cyan). The numbers $32$ and $64$ denote Input image sizes $32\times32$ and $64\times64$, respectively; v1 stands for DW Capsule whereas v2 stands for SC Capsule. The next number expresses the amount of ConvNets. The last number denotes whether the ConvNet is followed by a Max Pooling (1 if not, 2 if followed). All ConvNets in the primary layer have the kernel sizes vary from $9\times9$, $7\times7$, $5\times5$ to $3\times3$.}
  \label{fig:dw_vs_sc}
\end{figure*}

In this experiment, we perform experimental evaluations of DW Capsules against SC Capsules. First, we vary the size of Convolution's kernel including $9\times9$, $7\times7$, $5\times5$ and $3\times3$ using the Input image's size of $32\times32$. Then we add one more ConvNet in the first layer in the Capsule Networks's architecture as shown in Figure \ref{fig:asl_capsule}. We also provide two variations of Capsule Networks, one for scaling down the total number of model parameters by adding a Max Pooling after the second ConvNet (Mini version) and, another, which increases Input image sizes to $64\times64$ (Max version).

Figure \ref{fig:dw_vs_sc_b} and \ref{fig:dw_vs_sc_d} show that DW Capsules perform equivalently or even better on accuracy than SC Capsules. Additionally, SC Capsules seems to be unstable and fluctuating based on kernel's size. Figure~\ref{fig:dw_vs_sc_a} shows a similar trend when SC Capsules are more likely to volatile when a half of SC Capsules are outperformed by DW Capsules. Only in Figure \ref{fig:dw_vs_sc_c}, all SC Capsules achieve higher accuracy than that of DW Capsules. Though this can be a trade of between training epoch and training speed.
\subsection{Experiment 2: Deep Learning Models using MLP and LSTM}
\begin{figure*}[p]
  \centering
  \subfloat[MLP vs LSTM on Inception V3]{\includegraphics[width=0.475\textwidth]{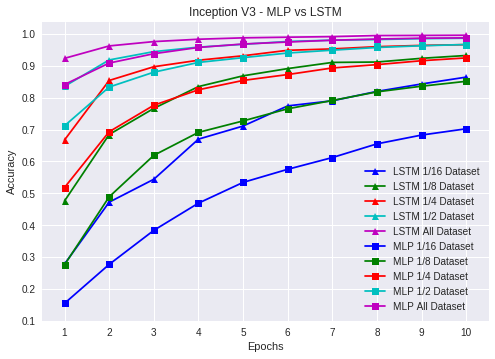}
  \label{fig:mlp_lstm_inceptionv3}}
  %
  \subfloat[MLP vs LSTM on MobileNet V1]{\includegraphics[width=0.475\textwidth]{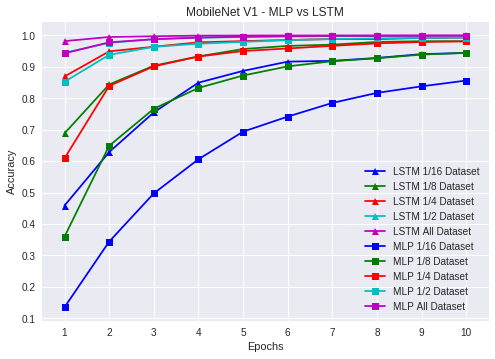}
  \label{fig:mlp_lstm_mobilenetv1}}

  \caption{Comparisons of MLP and LSTM on Deep Learning Models.}
  \label{fig:MLP_LSTM_TLModels}
\end{figure*}

In this section, we perform experiments of Deep Learning models using MLP and LSTM on variations of the ASL dataset including $\frac{1}{16}$, $\frac{1}{8}$, $\frac{1}{4}$, $\frac{1}{2}$ and the whole dataset.

Due to the limited number of pages allowed, we excerpt results from most of all Deep Learning models and preserve only one model for mobile devices and one model for powerful computers i.e. MobileNet V1 and Inception V3. We select MobileNet V1 because of its best accuracy and Inception V3 since the model is faster than DenseNet V201.

The Figure \ref{fig:mlp_lstm_inceptionv3} shows that Inception V3 Transfer Learning model when integrated with LSTM outperforms its version on MLP in all sets of data. Similarly, MobileNet V1 LSTM achieves a higher accuracy than MobileNet V1 MLP. Remarkably, MobileNet V1 performs better than Inception V3 on both MLP and LSTM versions even though the model is mainly built for much smaller devices.
\subsection{Capsule Networks vs Deep Learning Models on Model Size and Accuracy}
\begin{figure*}[!t]
\begin{center}
\includegraphics[width=0.95\textwidth]{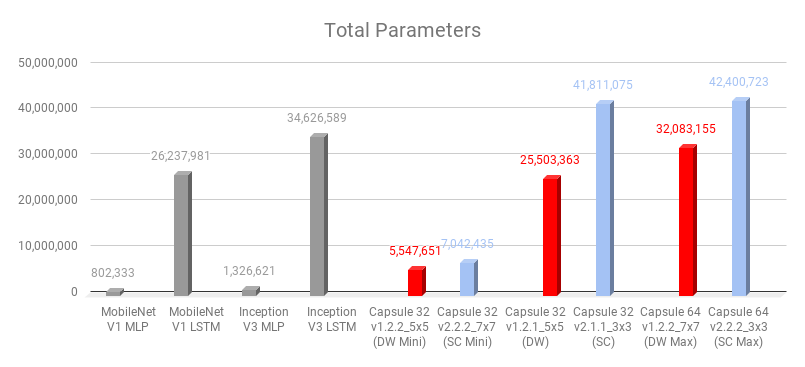}
\caption{Total Parameters of Capsule Networks and Deep Learning Models. The numbers $32$ and $64$ denote Input image sizes $32\times32$ and $64\times64$, respectively; v1 stands for DW Capsule whereas v2 stands for SC Capsule. The next number expresses the amount of ConvNets. The last number denotes whether the ConvNet is followed by a Max Pooling (1 if not, 2 if followed). All ConvNets in the primary layer have the kernel sizes vary from $9\times9$, $7\times7$, $5\times5$ to $3\times3$.}
\label{fig:total_parameters}
\end{center}
\end{figure*}
In this Section, we analyse Capsule Networks and Deep Learning Models with respect to the total size of these models as well as accuracy. We select the best DW Capsules and SC Capsules including: 1. Input image size $32\times32$ pixel followed by two ConvNets and a Max Pooling (Figure \ref{fig:dw_vs_sc_c}) 2. Input image size $32\times32$ pixel one for the best accuracy SC (Figure \ref{fig:dw_vs_sc_a}) and one for the best accuracy DW (Figure \ref{fig:dw_vs_sc_b}) 3. Input image size $64\times64$ pixel integrated with two ConvNets and one Max Pooling (Figure \ref{fig:dw_vs_sc_d}). These Capsule Networks are denoted as DW Mini, SC Mini, DW, SC, DW Max and SC Max, respectively.

Generally, we can see from Figure \ref{fig:total_parameters} that DW Capsules drastically decrease the models' size. More specifically, Capsule 32 DW Mini reduces the model size by $21\%$ while Capsule 64 DW Max shrinks $25\%$ of the total parameters. It can be noted that Capsule 32 DW reduces the number of parameters by $40\%$ which is more than Capsule 32 DW Mini and Max due to one more Convolutions.

In comparison between Capsule Networks and Deep Learning models, we can observe that MobileNet V1 MLP has a smallest model size followed by Inception V3 MLP. Additionally, Capsule 32 DW and Capsule 64 DW Max have smaller sizes than MobileNet LSTM and Inception V3 LSTM.

In terms of accuracy, MobileNet V1 LSTM outperforms all other models. In spite of that, Capsule 64 DW Max reaches the second position and performs better than MobileNet V1 MLP after $10$ epochs. In addition, Capsule 64 DW Max outperforms both versions of Inception V3 LSTM and MLP by a large extent. Though Capsule 64 DW has $40$ times larger model size than MobileNet V1 MLP and $20\%$ more than MobileNet V1 LSTM, its total parameter is less than that of Inception V3 LSTM.
\begin{figure*}[!t]
\begin{center}
\includegraphics[width=0.95\textwidth]{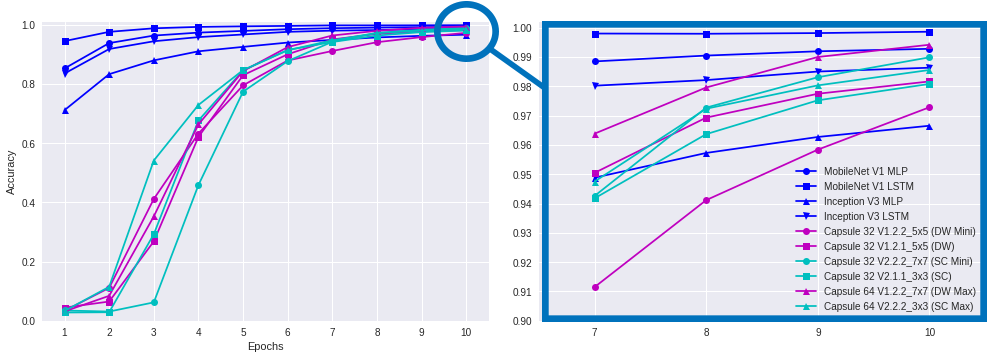}
\caption{Capsule Networks vs Deep Learning Models}
\label{fig:capsule_vs_trasnferlearning}
\end{center}
\end{figure*}

It can be noticed that Capsule 32 DW Mini though having the smallest size among all Capsule Networks but is outperformed by MobileNet V1 MLP with regards to accuracy. However, when we perform these experiments with more epochs, this gap can be eliminated as Capsule 32 DW Mini yields the accuracy of MobileNet V1 MLP ($0.9928$) after 17 epochs. A similar trend also occurs to Capsule 32 SC Mini as it reaches the accuracy of MobileNet V1 LSTM ($0.9986$) after 18 epochs (approximately 1h training on Tesla K80).

\section{Conclusions}
In this research, we first propose to replace Standard Convolution in Capsule Networks' Architecture with Depthwise Separable Convolution. Then we perform empirical comparisons of the best Capsule Networks with the best Deep Learning models.

The results show that our proposed DW Capsules remarkably decrease the size of models. Among the chosen Capsule Networks, the total parameters had shrunk roughly by an amount between $21\%$ -- $25\%$.

In terms of accuracy, Capsule 64 DW Max performs better than other Capsule models. Though Capsule 32 SC Mini can be a trade-off between the accuracy and the number of parameters.

In comparison with Deep Learning models, Capsule 32 DW Mini has a larger number of parameters, yet it achieves the accuracy of MobileNet V1 after a few more epochs. Meanwhile, Capsule 32 SC Mini can attain that of MobileNet V1 LSTM's accuracy with $3$ to $4$ times smaller in the number of parameters.

Capsule 32 DW Mini and 64 DW Max outperform Inception V3 MLP and LSTM on accuracy. Moreover, Capsule 64 DW Max occupies $5\%$ less the number of parameters than Inception V3 LSTM though Capsule 32 DW Mini has $4$ times larger size than Inception V3 MLP.


After a thorough literature search, we believe that this is the first work that proposes the integration of Depthwise Separable Convolution into Capsule Networks. Additionally, we provide empirical evaluations of the proposed Capsule Networks versus the best Deep Learning models.

In future work, we will apply the proposed Capsule Networks on  different datasets and will develop these networks for mobile platforms.
\label{sec:conclude}
\bibliographystyle{splncs04}
\bibliography{ibPRIA2019}
\end{document}